\documentclass[letterpaper, 10 pt, conference]{ieeeconf}  

\IEEEoverridecommandlockouts                              

\usepackage{graphicx}
\usepackage{subfigure}
\usepackage{amsmath} 
\usepackage{amssymb}  
\usepackage{gensymb}
\usepackage{cite}

\title{\LARGE \bf
LMBAO: A Landmark Map for Bundle Adjustment Odometry in LiDAR SLAM}

\author{Letian Zhang$^{1}$, Jinping Wang$^{1}$, Lu Jie$^{1}$, Nanjie Chen$^{1}$, Xiaojun Tan$^{1}$, Zhifei Duan$^{2}$ 
\thanks{*This work was supported by the Key-Research and Development Prgoram of Guangdong Province under Grand (2020B0909030005 and 2020B090921003). (Corresponding author:	Xiaojun Tan.)}
\thanks{$^{1}$Letian Zhang, Jinping Wang, Lu Jie, Nanjie Chen and Xiaojun Tan are with the School of Intelligent Systems Engineering, Sun Yat-sen University, Guangzhou 510006, China (e-mail: zhanglt33@mail2.sysu.edu.cn; wangjp29@mail2.sysu.edu.cn; jielu@mail2.sysu.edu.cn; chennj6@mail2.sysu.edu.cn; tanxj@mail.sysu.edu.cn).}
\thanks{$^{2}$Zhifei Duan is with the Automotive Technology Center, XPeng Inc., Guangzhou 510640, China (e-mail: duanzf@xiaopeng.com).}
}
\begin{document}

\maketitle
\thispagestyle{empty}
\pagestyle{empty}

\begin{abstract}
LiDAR odometry is one of the essential parts of LiDAR simultaneous localization and mapping (SLAM). However, existing LiDAR odometry tends to match a new scan simply iteratively with previous fixed-pose scans, gradually accumulating errors. Furthermore, as an effective joint optimization mechanism, bundle adjustment (BA) cannot be directly introduced into real-time odometry due to the intensive computation of large-scale global landmarks.
Therefore, this letter designs a new strategy named a landmark map for bundle adjustment odometry (LMBAO) in LiDAR SLAM to solve these problems. First, BA-based odometry is further developed with an active landmark maintenance strategy for a more accurate local registration and avoiding cumulative errors. Specifically, this paper keeps entire stable landmarks on the map instead of just their feature points in the sliding window and deletes the landmarks according to their active grade. Next, the sliding window length is reduced, and marginalization is performed to retain the scans outside the window but corresponding to active landmarks on the map, greatly simplifying the computation and improving the real-time properties.
In addition, experiments on three challenging datasets show that our algorithm achieves real-time performance in outdoor driving and outperforms state-of-the-art LiDAR SLAM algorithms, including Lego-LOAM and VLOM.

\end{abstract}

\section{INTRODUCTION}
Simultaneous localization and mapping (SLAM) is one of the most fundamental problems in robotic applications, especially autonomous navigation. 
With Visual-based and LiDAR-based sensing, great efforts have been made to achieve highly accurate real-time localization.
LiDAR is known as a reliable, illumination-insensitive sensor that can detect the fine details of an environment in a large area. Therefore, this letter focuses on an implementation method for LiDAR SLAM in outdoor driving.

The classic LiDAR SLAM framework \cite{b1,b2,b3,b4,v11,v12,b5,b6} registers LiDAR scans incrementally and estimates only the pose of the current frame each time. In fact, the current frame can also improve the estimates of the historical frames, which in turn improves the estimate of the current frame. The introduction of bundle adjustment (BA), which jointly optimizes the pose of multiple frames, avoids error accumulation, and effectively lowers the drift in LiDAR SLAM \cite{b7,b8,b9}.

To provide enough information for optimization, the length of the BA sliding window of map refinement is generally large (set to 20 in \cite{b7}). However, when BA is used for odometry, the computational cost is extremely limited due to the high real-time requirements, which means that the length of the sliding window is restricted (set to 8 in \cite{b9}). Thus, only information from recent frames is available in the map, and previous constraints from earlier frames cannot be used. The insufficient constraints for BA have implications for the applicability to odometry. Large errors occur when driving with obvious pose changes or encountering large obstacles.

\begin{figure}[tp]
	\centering
	\includegraphics[width=0.99\linewidth]{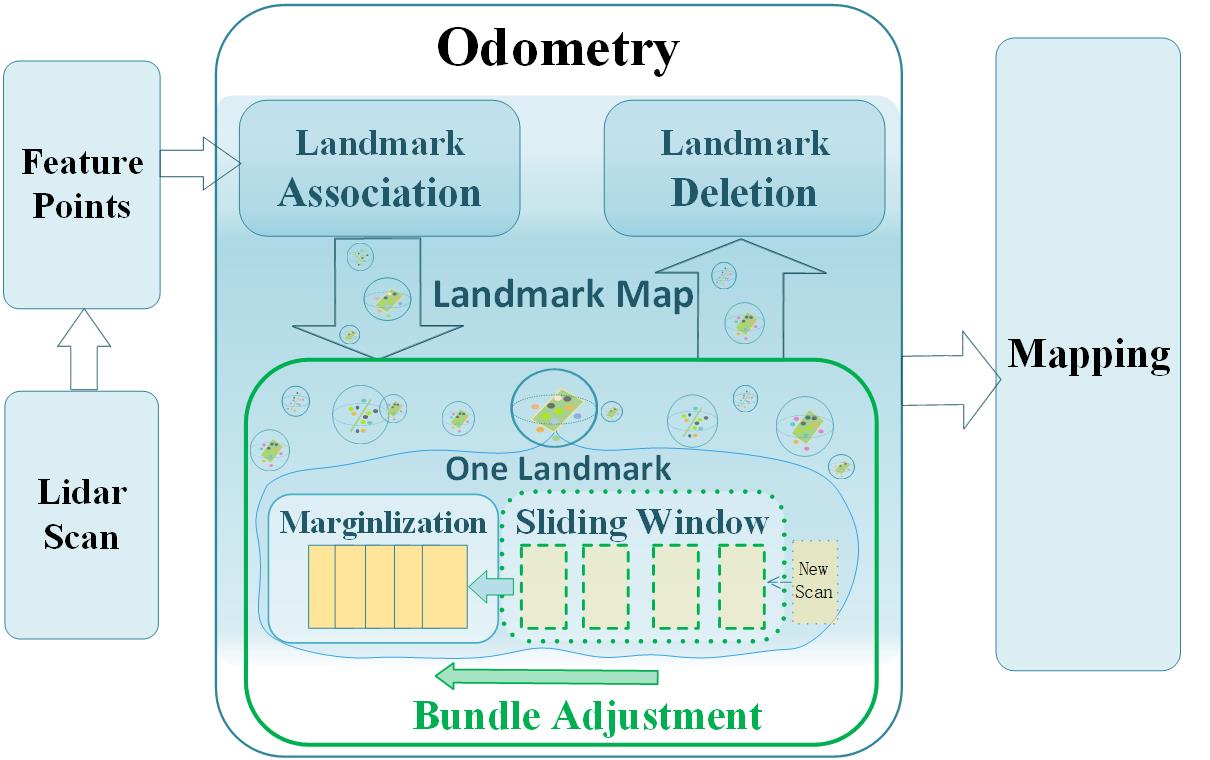}
	\caption{Our work proposes a BA odometry using active landmark map maintenance including landmark association and landmark deletion. We feed BA with landmarks from all scans and optimize only the pose of the scans inside the sliding window (in the green box).}
	\label{fig:system}
\end{figure}
An overview of our system is shown in Fig. \ref{fig:system}.
Our work further develops the BA odometry for LiDAR SLAM by using an active landmark maintenance strategy to separate the map from the sliding window (whose length is fixed at 4 in LMBAO). The map is divided into a sliding window part involved in the joint optimization of pose and velocity, and a marginal part outside the window providing sufficient prior constraints.
The main contributions of our work can be summarized as follows:
\begin{itemize}
\item We propose a strategy for maintaining local landmark maps that is independent of the sliding window and aims to add selected prior constraints to the map.
With an extremely limited capacity, the local landmark map is suitable for BA odometry to avoid cumulative errors.
\item We develop an observation count to control the lifetime of landmarks and a center drift to discard unstable landmarks, successfully avoiding interference from unanticipated occlusion and dynamic objects.
\item To speed up the computation, we reduce the number of scans to be optimized in the sliding window and perform marginalization in combination with our landmark map to fully exploit prior constraints. By using an incremental covariance matrix, marginalization retains the scans related to active landmarks but outside the window.
\end{itemize}

Experimental results show that our algorithm achieves real-time performance and outperforms state-of-the-art LiDAR SLAM algorithms, including Lego-LOAM and VLOM. 

\section{RELATED WORK}
The essential task of LiDAR SLAM is matching scans. 
Iterative closest points (ICP) \cite{b1} is a classical method for determining the correlation transformation of two raw point clouds by directly matching all points through laborious computation. Several ICP variants \cite{b2,b3,b4} have been proposed to increase efficiency. In \cite{b3}, generalized-ICP (GICP) attempts to match local planar patches from both scans and integrates point-to-plane distances to ICP. 
Feature-based matching methods effectively reduce computational costs by matching only feature points extracted from raw point clouds. In the pioneering LOAM work \cite{b5}, planes and edges are extracted based on local smoothness analysis and matched with features from a previous scan, enabling real-time scan-to-scan registration. In parallel, a lower frequency algorithm registers the scan on the map to refine its pose.   
Lego-LOAM \cite{b6} adds ground segmentation to improve the accuracy of feature extraction of LOAM \cite{b5}. VLOM \cite{b9} introduces spherical projection \cite{b12,b13} to obtain the spatial structure of point clouds, and extracts feature points from spherical images in a novel way, achieving accurate correlation of lines and planes.
Some deep-learning based approaches \cite{b10,b11} are also used to extract 3D features from sparse point clouds. 

\textbf{Sliding Window:} 
A sliding window is introduced in the odometry to register the new scan on a local map rather than just on the previous scan. 
Forgoing the global map used in the mapping step \cite{b5}, LIO-SAM \cite{b14} implements a sliding window to create a local feature map, that allows real-time registration of a new keyframe to a  set of prior subkeyframes. 
Similarly, Ye {\em et al.} build a local map using a sliding window instead of a single scan \cite{b15}, so that the points are dense enough to accurately estimate the relative scan pose.
LIC-Fusion2.0 \cite{b16} tracks planars across multiple scans within a sliding window by comparing the point-plane distance and parallelity of normal vectors, which improves the robustness of their scan-to-scan matching model \cite{b17}. However, it is only verified indoors \cite{b16}.
Although a sliding window is used to store the local prior constraints, the above methods \cite{b14,b15,b16} only use it to compute the relative pose of the new frame. Due to computational complexity, these algorithms do not fully utilize the stored concurrent constraints to optimize all frame poses within the window. 
Our framework optimizes scans using the BA algorithm with a size-limited and constraint-sufficient landmark map to improve efficiency. 

\textbf{Local Optimization with Bundle Adjustment:} 
Following the visual SLAM \cite{b26}, a local BA is performed over a sliding window of LiDAR scans to effectively reduce drift. Zheng {\em et al.} developed a theoretical formulation of LiDAR BA that includes the first and second order derivatives of the cost function \cite{b7}, enabling accelerated optimization with the Levenberg-Marquardt (LM) method \cite{b18}. BALM \cite{b7} integrated a local BA in \cite{b5} as a back-end to refine the incrementally generated map. Although \cite{b5} used a voxel map and marginalization \cite{b19}, the complexity of the calculations limited its application to high-speed outdoor driving. $\pi$-LSAM \cite{b8} proposes a local-global plane association method to estimate the relative transformation of the new scan in the front-end odometry, and introduces the plane adjustment (PA) \cite{b25} into back-end scan-to-map matching. 
In subsequent work \cite{b20}, Zhou {\em et al.} simplify the relative pose estimation process in odometry and add the integrated cost matrix (ICM) to the local PA process in the back-end to make use of the prior plane constraints. However, \cite{b20} focuses on indoor application scenarios and only considers the incremental process, where the size of the map and the accumulated historical information are limited, making it difficult to migrate to outdoor areas. Unfortunately, both \cite{b8} and \cite{b20} do not make use of edge features.

\textbf{Global Feature Map in Local BA:} 
The use of BA described above is still limited to the mapping step, which requires a separate, previously performed scan-to-scan estimation step as odometry.
Instead of performing scan-to-scan alignment, VLOM \cite{b9} creatively predicts the pose of a new frame immediately from the previous frame with a constant motion model. Feature points extracted within the frame can be directly associated with global features, and used in local BA to optimize velocities and poses. 
To achieve high real-time odometry performance, the size of its global feature map, which is entirely involved in the local BA, is extremely limited. Unfortunately, similar to BALM \cite{b7} and $\pi$-LSAM \cite{b8}, VLOM maintains its global feature map simply depending on the sliding window. This coarse maintenance method without filters results in very limited prior constraints available in the map.
For this reason, we improve the map maintenance strategy in \cite{b9} to fully exploit the prior constraints and introduce an appropriate marginalization mechanism on the sliding window to reduce the computational cost. 

\section{A Percise Landmark Map}

\subsection{Motion Compensation and Prediction}
We define the $k$th LiDAR scan as $S_{k}$ and its acquisition time within $\left[ t_{k},t_{k+1}\right)$. $t_{k}$ is equal to $t_{k}^0$, the timestamp of the first point in $S_{k}$ and $t_{k+1}$ is the start time of $S_{k+1}$. The pose in the coordinate system of the current scan $\left\lbrace\mathbf{ C_{k}} \right\rbrace$ is regarded as $^C\!{P}_{k}$
, which is equal to $^C\!{p}_{k}^0$, the position of the first point in $S_{k}$. 
We transform $S_{k}$ from $\left\lbrace\mathbf{ C_{k}} \right\rbrace$ to the world coordinate system $\left\lbrace\mathbf{W} \right\rbrace$ with the transformation $T_{k}\!\in\!SE(3)$.

\begin{figure}[tp]
	\centering
	\includegraphics[width=0.99\linewidth]{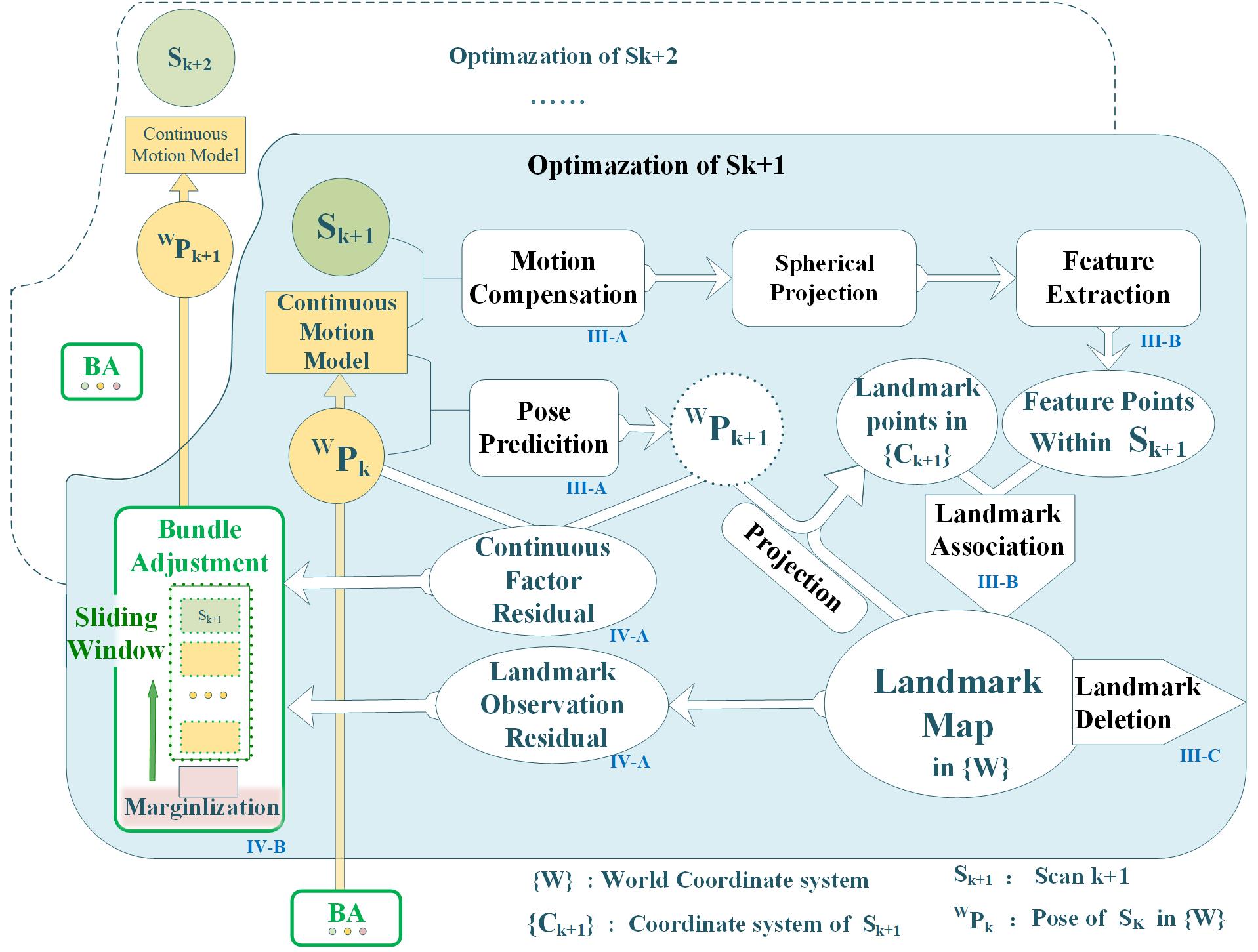}
	\caption{The BA odometry process. Using the continuous motion model, feature points are compensated and predicted to construct landmarks. Each landmark stored in the map contains all points and forms a residual that participates in the local BA optimization. Marginalization is used to speed up the calculation.}
	\label{fig:odo}
\end{figure}

\textbf{Motion Compensation:}
To compensate for the distortion of the LiDAR ego-motion, we assume that the LiDAR is constantly rotating at angular velocity $\omega\!\in\mathbb{R}^3$, and moving at a linear velocity $v\!\in\!\mathbb{R}^3$, as \cite{b9}. Both $\omega$ and $v$ are defined  in the current scan coordinate system. We use a rotation vector to represent $\omega$. The skew matrix of $\omega$ is defined as $[\omega]_{\times}\! \in \! so(3)$. Then the exponential map $ exp: so(3)\!\rightarrow \!SO(3)$ has the following form:
\begin{equation}\label{wx} 
exp(\omega) = \textbf{I} +\frac{sin(\Vert \omega \Vert)}{\Vert \omega \Vert}{[\omega]_{\times}} + \frac{1-cos(\Vert \omega \Vert)}{\Vert \omega \Vert^2}{[\omega]_{\times}^2} \ .
\end{equation}
For any point $i$ in scan $S_{k}$, the recorded time is $t_{k}^i$, and its pose in raw point cloud $\left\lbrace\mathbf{C_{k}} \right\rbrace$ is $^{C}p_{k}^i$.  The pose after motion compensation is $^{C}\tilde{p}_{k}^i$ as \eqref{C}.
\begin{equation}\label{C} 
^{C}\!\tilde{p}_{k}^i=exp(({t_{k}^i}-t_{k})\omega_{k}) \ ^{C}\!p_{k}^i+({t_{k}^i}-t_{k})v_{k} \ .
\end{equation}

The transformation $T_{k}$ of $S_{k}$ is represented by rotation matrix $\textbf{\textit{R}}$ and translation vector $\textbf{\textit{t}}$, which are defined in the world coordinate system $\left\lbrace\mathbf{W} \right\rbrace$. For point $i$, the world position with motion compensation $^{W}\!\tilde{p_{k}^i}$ is
\begin{equation}\label{G} 
^{W}\!\tilde{p}_{k}^i=T_{k} \ ^{C}\!\tilde{p}_{k}^i  \ .
\end{equation}

\textbf{Pose Prediction of New Scan:}
Since $t_{k+1}$ is the starting time of $S_{k+1}$, we can  predict the pose of next scan $^C\!\tilde{P}_{k+1}$ relative to $^C\!{P}_{k}$ according to \eqref{C}. Furthermore, the transformation matrix $\tilde{T}_{k+1}$ of $S_{k+1}$ as well as ${v}_{k+1}$ and ${\omega}_{k+1}$ can be predicted solely on the basis of $S_{k}$ and the continuous model
\begin{equation}\label{xk+1} 
\left\{
\begin{aligned}
\tilde{T}_{k+1} &=T_{k} \! 
\left[\!\begin{array}{cc}
exp(({t_{k+1}}-t_{k})\omega_{k})&({t_{k+1}}-t_{k})v_{k}  \\
0 & 1 
\end{array}\!\right]\\
\tilde{v}_{k+1} &=exp(-(t_{k+1}-t_{k})\omega_{k})v_{k} \ \ \ \ \ \ \ \ \ \ \ \ \ \ \ \ \ \ \ \ \ \ . \\
\tilde{\omega}_{k+1} &=exp(-(t_{k+1}-t_{k})\omega_{k})\omega_{k} 
\end{aligned}
\right. 
\end{equation}

With state variable $\tilde{x}_{k+1}=\left\lbrace \tilde{T}_{k+1},\tilde{v}_{k+1},\tilde{\omega}_{k+1}\right\rbrace$, \eqref{C} is applied again to compensate  the points of $S_{k+1}$ in $\left\lbrace\mathbf{C_{k+1}} \right\rbrace$. Therefore, the distortion can be directly eliminated in the subsequent pose optimization steps. Using $\tilde{T}_{k+1}$, the compensated feature points of $S_{k+1}$ can be accurately associated with points of previous scans in $\left\lbrace\mathbf{W} \right\rbrace$, as shown in Fig. \ref{fig:odo}.

\subsection{Landmarks Construction with Feature Points}
The process of extracting feature points and constructing landmarks is similar to the method used in VLOM \cite{b9}. Once a new scan arrives, feature points are first extracted from this single scan. 
Then these feature points are associated across multiple scans to construct global landmarks that are stored and maintained in maps.

\textbf{Feature Point Extraction from a Single Scan:}
For each point $i$ of the $S_{k}$ obtained from rotary LiDAR, we remove the distortion using \eqref{C}, so $^C\!\tilde{p}_{k}^i=[x,y,z]$. The spherical projection coordinate $[u,v]^T$ is projected as
\begin{equation}\label{spherical} 
\left[
\begin{array}{c}
u  \\
v  
\end{array}
\right]
=
\left[
\begin{array}{c}
\alpha(t_{k}^i-t_{k}) \\
r  
\end{array}
\right] \ ,
\end{equation}
where $r=\sqrt{x^2+y^2+z^2}$ is the distance from point $i$ to the origin and $\alpha$ is a preset parameter according to the specific laser resolution.
Then a spherical 2D image is obtained.

The smoothness score of a point is calculated using the neighborhood of each point in the spherical image. Based on the smoothness score, feature points that may be located on one plane/edge are selected. Then the selected feature points are clustered by the breadth first search procedure and marked with the feature ID of the cluster to which they belong
The details can be found in \cite{b9}.

\textbf{Landmark Association Across  Multiple Scans:}
We use the newly extracted feature points with their feature ID to construct global landmarks. Landmark $L$ is a spherical region with feature points and is represented by center $c_{L}$ and radius $r$. Landmarks have their own category that identifies them as edges or planes.

When the new scan $S_{k+1}$ arrives, a pose prediction and motion compensation are described in Section III-A.
All landmarks in the map are also projected into $\left\lbrace\mathbf{C_{k+1}} \right\rbrace$ with the predicted $\tilde{T}_{k+1}$ to accurately associate these feature points.
The core idea of feature association is that a feature point in a new scan approaching a landmark is considered an observation point of the landmark and added to the landmark. Landmarks in the map continuously collect points from different scans, which is called landmark tracking. To avoid redundancy, new landmarks are created only from feature points that are not tracked by existing landmarks.
When tracking and creating landmarks, we save the feature points using the position from the raw point cloud, not  the position after compensation.


\subsection{Landmark Deletion from Map}
To reduce the computational burden, we simultaneously delete landmarks to limit the number of points in the map.
Unlike \cite{b7} and \cite{b9}, which rely on the sliding window,
the feature points of one scan are deleted when it slides out of the window.
We adopt a more active map maintenance strategy that uses landmark active degree to distinguish landmarks and fully preserves feature points from active landmarks.
The landmark active degree consists of three parts: the observation count $O_{L}$, the drift of the center point $c_{L}$ in optimization and the feature points number $N^L$.

\textbf{Observation Count $O_{L}$:}
When a new landmark is created, the observation count $O$ is initialized to a fixed value (4 in this letter). Whenever a new scan arrives and the landmark successfully tracks the feature points in that scan, the count is increased by one; otherwise, the count is decreased by one. When $O_{L}$ decreases to 0, the landmark is deleted from the map. Even if the landmark still has feature points that participate in the sliding window optimization, they will be deleted directly. In this way, landmarks that can be continuously tracked are not immediately discarded by the sliding window when they encounter unexpected obstructions, so they can always participate in the optimization.

\textbf{Drift of Landmark Center $c_{L}$:}
When creating a landmark, the center point $c_{L}$ is computed using the position of its feature points in $\left\lbrace\mathbf{C_{k+1}} \right\rbrace$ and its global projection is saved as $c_{L}^0$ using the predicted $\tilde{T}_{k+1}$. The pose ${^W\!{P}_{k+1}}$ of this scan ${S}_{k+1}$ is constantly updated as it participates in sliding window optimization. When the scan $S_{k+1}$ optimizes the new pose ${^W\!{P}_{k+1}}^{'}$ in the sliding window, the global projection of $c_{L}$ is recomputed as $c_{L}^{'}$. The drift of $c_{L}$ is given by the distance between $c_{L}^0$ and $c_{L}^{'}$. If the drift exceeds $3r$, the feature points tracked by the landmark in the next scan are far from the feature points captured when the landmark was created. Then, the landmark will be deleted as it is an unstable landmark with large drift.

\textbf{Feature points number $N^L$:}
To reduce the influence of dynamic objects, which are usually small in size, we treat some small plane landmarks as unstable landmarks whose feature points are still below the threshold ($N^L < 86$) even after continuously tracking multiple scans, and remove them directly from the map. 

These three measures above effectively reduce the interference from random obstacles and the existence time of unstable landmarks.

\section{Bundle Adjustment Odometry with Marginalization}
\subsection{Construction of Residuals and Optimization Function}
\textbf{Landmark Observation Residual:} 
Each landmark contributes an observation residual to the BA optimization - a global pose optimization of the LiDAR scan in $\left\lbrace\mathbf{W} \right\rbrace$.
Since the landmark stores the points from the raw point cloud of scans and the pose of each scan is constantly updated during the optimization process, these points must be reprojected into $\left\lbrace\mathbf{W} \right\rbrace$.
Motion compensation is also performed each time before reprojection as in Section III-A.
 
In the world coordinate system $\left\lbrace\mathbf{W} \right\rbrace$, the covariance matrix ${Cov}$ of all points of landmark L is 
\begin{equation}\label{COv1} 
\left\{
\begin{aligned}
{^W\!\bar{p}_{L}} &=\dfrac{1}{N^L}\sum_{k=1}^{K}\sum_{j=1}^{N_{k}^L}{^W\!\tilde{p}_{k}}^j \\
Cov &=\dfrac{1}{N^L}\sum_{k=1}^{K}\sum_{j=1}^{N_{k}^L}({^W\!\tilde{p}_{k}}^j-{^W\!\bar{p}_{L}} ){({^W\!\tilde{p}_{k}}^j-{^W\!\bar{p}_{L}})}^T
\end{aligned}
\right. .
\end{equation}

For plane landmarks, we construct the loss function $\varepsilon_{\pi}$ corresponding to the distance of all feature points from the plane, as follows:
\begin{equation}\label{loss2p} 
\varepsilon_{\pi}^2=\sum_{k=1}^{N^L}{\Vert{n_{\pi}^T(^W\!p_{k}-{^W\!p_{\pi}})}\Vert}_{2}^2 \ .
\end{equation} 

In the above equation, the planar parameters (normal vector $n_{\pi}$, planar point $^W\!p_{\pi}$) must jointly participate in the optimization. To further simplify the equation, the optimal planar parameters are calculated in advance and eliminated before optimization.

BALM proved that in the optimal plane, the final loss is equivalent to the solution of the minimal eigenvalue of the point set. Therefore, \eqref{loss2p} of planar loss can be simplified to
\begin{equation}\label{lossp} 
\varepsilon_{\pi}=\sqrt{\lambda_{1}(Cov)} \ ,
\end{equation}
where $\lambda_{i}(Cov)$ represents the $i_{th}$ smallest eigenvalue of matrix $Cov$.

Similarly, for edge landmarks, the loss function $\varepsilon_{e}$ is constructed based on the  point-line distance
\begin{equation}\label{loss2e} 
\varepsilon_{e}^2=\sum_{k=1}^{N^L}{\Vert{(I-n_{e}{n_{e}}^T)(^W\!p_{k}-{^W\!p_{e}})\Vert}_{2}^2} \ .
\end{equation}

With reference to BALM, the final edge loss function can be simplified to
\begin{equation}\label{losse} 
\varepsilon_{e}=\sqrt{\lambda_{2}(Cov)+\lambda_{3}(Cov)} \ .
\end{equation}

\textbf{Continuous Factor Residual:} 
The continuous motion model requires a certain smoothness of LiDAR motion, which is satisfied in most cases. However, the linear velocity and angular velocity that need to be optimized are added to each scan and usually cause divergence. To better constrain the motion state, we add a continuous factor with respect to the preintegration factor IMU, whose residuals are defined as follows:
\begin{equation}\label{imu} 
\!r_{s}(x_{k},x_{k+1}\!)\!=\!
\left[
\begin{aligned}
&R_{k}(t_{k+1}-t_{k})v_{k}+p_{k}-p_{k}       \\
&\log(\exp((s_{k+1}\!-\!s_{k})\omega_{k})R_{k}/R_{k+1})\\
&R_{k}\omega_{k}-R_{k+1}\omega_{k+1}
\end{aligned} 
\right]\! .
\end{equation} 

\textbf{Optimization Function:} 
Global landmarks are the result of the association of feature points across multiple scans, so the position and survival time of each landmark are different. To standardize the format, each landmark is assumed to contain a total of $N^L$ points from all $K$ historical scans, and the number of points tracked by the scan $S_{k}$ is $N_{k}^L$. If there are no feature points in the landmark that belongs to this scan, $N_{k}^L=0$. Thus, if we use the feature points in the landmarks in the map to optimize the scan pose, we need the state vectors of all historical scans.

Let $\mathbb{\chi}\!=\!\left\lbrace x_{1}, x_{2}, x_{3},...,x_{K}\right\rbrace\!$ represent the set of state vectors of all historical scans, and $\mathbb{M}_{\pi}$ and $\mathbb{M}_{e}$ represent the set of all planar landmarks and all edge landmarks in the current map, respectively. The final optimization problem can be expressed as the sum of landmark observation residuals and smoothing residuals
\begin{equation}\label{argmin1}
\mathop{\arg\min}\limits_{\chi}\sum_{e\in{\mathbb{M}_{e}}}\rho({\Vert r_{L}^e (\chi)\Vert}^2)+\sum_{\pi\in{\mathbb{M}_{\pi}}}\rho({\Vert r_{L}^\pi (\chi)\Vert}^2)  $$ \\
$$ +\sum_{k=1}^{K-1}\Vert {r_{s}(x_{k},x_{k+1})\Vert}^2  \ ,  
\end{equation}
where Huber loss $\rho$ is defined as follows:

\begin{equation}\label{loss} 
\rho(s)= 
\left\lbrace 
\begin{aligned}
&s  ,&s <= 1      \\
&2\sqrt{s}-1 ,&s>1
\end{aligned} 
\right. \! .
\end{equation}

For each landmark observation residual $r_{L}^\pi$ or $r_{L}^e$,
the calculation process can be divided into two parts: calculation of covariance matrix $Cov$ and eigenvalue decomposition. The calculation of $Cov$ depends on all feature points of the landmark, which includes the pose of all historical scans. When the optimization formula is used directly, it is the bundle adjustment of all scans, and the pose of all scans is optimized in each iteration. Although our new map maintenance strategy can limit the number of landmarks in the map, the cost of the optimization formula increases over time and eventually becomes unacceptable. 
Following the visual SLAM \cite{b26}, this letter uses the sliding window mechanism to limit the number of historical scans to be optimized. The landmark information is fully stored in the residuals by the marginalization strategy, which provides a good prior for the subsequent optimization.

\subsection{Sliding Window and Marginalization}
The key problem to be solved by the marginalization is the calculation of the landmark observation residuals  when a scan slides out of the window. Suppose the size of the current sliding window is $n$. By decomposing the $Cov$ computational process, this letter rewrites \eqref{COv1} to describe the sliding window mechanism and marginalization strategy as follows:
\begin{equation}\label{cov2}
\begin{split}
Cov\!&=\!{\small \dfrac{1}{N^L}}\!\sum_{k=1}^{K}\!\underbrace{\!\sum_{j=1}^{N_{k}^L}\!{^W\!\tilde{p}_{k}^j}{(^W\!\tilde{p}_{k}^j)\!^T}}_{\tilde{O}_{k}}\!-\!{\small \dfrac{1}{N^L}}(\!\sum_{k=1}^{K}\!\underbrace{\!\sum_{j=1}^{N_{k}^L}\!{^W\!\tilde{p}_{k}^j}\!}_{\tilde{S}_{k}})(\sum_{k=1}^{K}\!\underbrace{\!\sum_{j=1}^{N_{k}^L}\!{^W\!\tilde{p}_{k}^j}\!)\!^T}_{\tilde{S}_{k}} \\
&=\dfrac{1}{N^L}\sum_{k=1}^{K}{\tilde{O}_{k}} -\dfrac{1}{N^L}(\sum_{k=1}^{K}{\tilde{S}_{k}})(\sum_{k=1}^{K}{\tilde{S}_{k}})^T    \\
&=\dfrac{1}{N^L}(\underbrace{\sum_{k=1}^{K-n}{\tilde{O}_{k}}}_{{O}_{marg}}+\underbrace{\sum_{k=K-n+1}^{K}{\tilde{O}_{k}}}_{\tilde{O}_{win}}) -\dfrac{1}{(N^L)^2}(\underbrace{\sum_{k=1}^{K-n}{\tilde{S}_{k}}}_{{S}_{marg}}  \\
&+\underbrace{\sum_{k=K-n+1}^{K}{\tilde{S}_{k}}}_{\tilde{S}_{win}})(\underbrace{\sum_{k=1}^{K-n}{\tilde{S}_{k}}}_{{S}_{marg}}+\underbrace{\sum_{k=K-n+1}^{K}{\tilde{S}_{k}}}_{\tilde{S}_{win}})^T  \ .
\end{split} 
\end{equation}

In \eqref{cov2}, we divide the calculation of $Cov$ into the marginal part ${O}_{marg}$, ${S}_{marg}$ and the sliding window part $\tilde{O}_{win},\tilde{S}_{win}$:
\begin{equation}\label{cov3}
\begin{split}
&Cov=\dfrac{1}{N^L}({O}_{marg}+\tilde{O}_{win})- \\ \dfrac{1}{(N^L)^2}&({S}_{marg}+\tilde{S}_{win})({S}_{marg}+\tilde{S}_{win})^T  
\end{split} \ .
\end{equation}

In the sliding window mechanism, only the state variables of the scans in the sliding window $\chi_{win}$ need to be optimized, while the poses of the scans outside the sliding window $\chi_{marg}$ is fixed. That is, the state variable $x_{marg}=\left\lbrace{T_{marg},v_{marg},\omega_{marg}}\right\rbrace$ of the marginalized scan is fixed in the subsequent optimization process. Using the transformation \eqref{G}, the positions of all points in $\chi_{marg}$ under the world coordinates are also determined.
According to the \eqref{cov2}, once $\chi_{marg}$ is fixed, ${O}_{marg}$ and ${S}_{marg}$ are also fixed. Thus ${O}_{marg}$ and ${S}_{marg}$ can be calculated in advance before optimization and fixed in each iterative optimization process.

The changes in ${O}_{marg}$ and ${S}_{marg}$ at each iteration correspond only to the scan sliding out of the window.
They can be accumulated, so it is not necessary to compute from the origin.
When the next scan $S_{K+1}$ enters the window, we can calculate ${O_{marg}^{K+1}}$ and ${S_{marg}^{K+1}}$ before optimization as in \eqref{cmarg}, depending on the pose of $S_{K-n+1}$ which just slides out.

\begin{equation}\label{cmarg}
\left\{
\begin{aligned}
{O_{marg}^{K+1}}= {O}_{marg}^K+\tilde{O}_{K-n+1}\\
{S_{marg}^{K+1}}= {S}_{marg}^K+\tilde{S}_{K-n+1}
\end{aligned} \ .
\right.
\end{equation}

$\tilde{O}_{win}$ and $\tilde{S}_{win}$ contain all scans in the sliding window, and the results must be recalculated at each iteration. When a new scan $S_{K+1}$ is added to the sliding window, its initial state vector $x_{K+1}$ is predicted using \eqref{xk+1} from the continuous motion model corresponding to $x_{K}$ of the previous scan. Then, optimization continues using the optimized pose. Until the scan slides out of the window, the last optimized pose is the fixed pose.

With such a sliding window and a marginalization mechanism, the number of scans that need to be optimized each time is greatly reduced. Let $\chi\!=\!\left\lbrace\chi_{marg},\chi_{win}\right\rbrace$, $\chi_{win}=\left\lbrace x_{K-n+1},...,x_{K}\right\rbrace$. The final optimization equation can be rewritten as
\begin{equation}\label{argmin2}
\mathop{\arg\min}\limits_{\chi_{win}}\sum_{e\in{\mathbb{M}_{e}}}\rho({\Vert{r_{L}^e}(\chi_{win})\Vert}^2)+  \sum_{\pi\in{\mathbb{M}_{\pi}}}\rho{\Vert{r_{L}^\pi}(\chi_{win})\Vert}^2  $$\\ 
 $$+\sum_{k=1}^{K-1}\Vert {r_{s}(x_{k},x_{k+1})\Vert}^2  \ . 
\end{equation}

Finally, the L-M method is used to solve \eqref{argmin2}. In each iteration, the Jacobi matrix of each residual is computed and used to obtain the estimate of the state vector of each scan by using the Ceres Slover library \cite{b24}.

\begin{table*}[tp]
	\caption{RMSE of the Mapping APE $(m)$ on the Public Datasets}
		\centering
		\renewcommand\arraystretch{1.2}
		\begin{tabular}{c|cccccccccccc}
			\hline
			Dataset&Utbm1 &Utbm2 &Utbm3& Utbm4 &Utbm5 &Utbm6 &Uluk1 &Uluk2 &Uluk3 &Uluk4 &Uluk5   \\
			\hline
			Length(km) & 5.207  &5.113 & 5.000&5.036&6.502 &5.140 &0.556 &0.741&0.235 &0.612 &0.552\\
			\hline\hline
			Lego-LOAM &155.6  &22.54   &14.71  &15.40 &26.87 &12.90 &3.106&3.606&3.943 &2.619 &3.440  \\
			VLOM     &22.51 & 23.96 & 23.30  &21.60 &24.33 &19.05 &2.811&2.921  &3.589 &2.482 &3.111  \\
			Ours     &\textbf{14.09}	&\textbf{12.60} &\textbf{4.244} &\textbf{6.963} &\textbf{7.796} &\textbf{12.49}&\textbf{2.605}  &\textbf{2.463}  &\textbf{3.027}  &\textbf{2.434} & \textbf{2.760}  \\
			\hline
		\end{tabular}
	\label{public dataset}
\end{table*}

\section{LiDAR Mapping}
Based on the mapping part of Lego-LOAM \cite{b6}, we developed a mapping algorithm that fully exploits the prior information stored in the odometry step. To improve the pose transformation, we match the landmarks $\left\lbrace L_{p}^t, L_{e}^t\right\rbrace $ updated from the new frame with the surrounding point cloud map $S^t$. Then, the L-M method is used again to obtain the final pose.

In our method, the features extracted in a single image are no longer stored directly, but the landmarks are updated based on the features. Landmarks that have not yet been updated are stored in the environment map $M^t$ as a priori information. This means that $M^t$ has stored all previous landmarks, so $M^t=\left\lbrace \left\lbrace L_{p}^1, L_{e}^1\right\rbrace,..., \left\lbrace L_{p}^t, L_{e}^t\right\rbrace  \right\rbrace $. Each landmark set is associated with the odometry pose of the frame, that is, the odometry pose of the frame is used to compensate for the distortion of the points in the landmark set by the continuous motion model.

The method of creating the global environment map $S^t$ from  $M^t$ can refer to \cite{b6}, including two methods. One is local map matching. The other is updating the pose estimated by using the factor map pose optimization framework. 

\section{EXPERIMENTS}
We conducted a series of experiments to verify the effectiveness of the proposed framework. The experiments were run with the Robot Operating System (ROS) on a laptop equipped with an AMD Ryzen7 5800H CPU and 16 GB RAM. We compare our algorithm with state-of-the-art LiDAR SLAM algorithms, including Lego-LOAM \cite{b6} and VLOM \cite{b9}. Then, we perform the ablation experiments to validate the improvement of LMBAO in odometry. Finally, we give the running time of our algorithm. In the following experiments, we run all compared algorithms 5 times on each dataset. The root mean square error (RMSE) of the absolute translational error (ATE) \cite{b21} is used as the evaluation index. 

\begin{figure}[tp]
	\centering
	\subfigure[Utbm2]
	{
		\begin{minipage}[b]{.46\linewidth}
			\centering
			\includegraphics[scale=0.55]{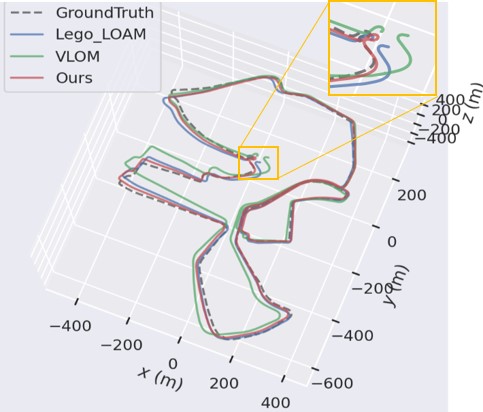}
		\end{minipage}
	}
	\subfigure[Utbm3]
	{
		\begin{minipage}[b]{.46\linewidth}
			\centering
			\includegraphics[scale=0.55]{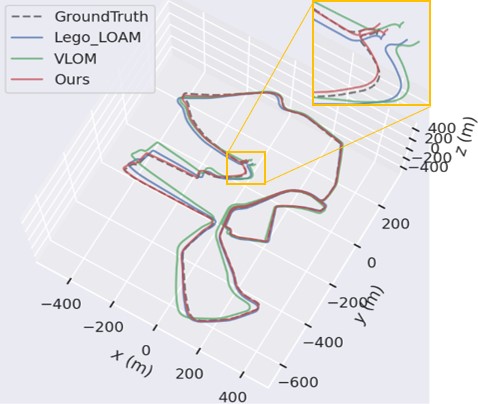}
		\end{minipage}
	}
	\subfigure[Utbm4]
	{
		\begin{minipage}[b]{.46\linewidth}
			\centering
			\includegraphics[scale=0.55]{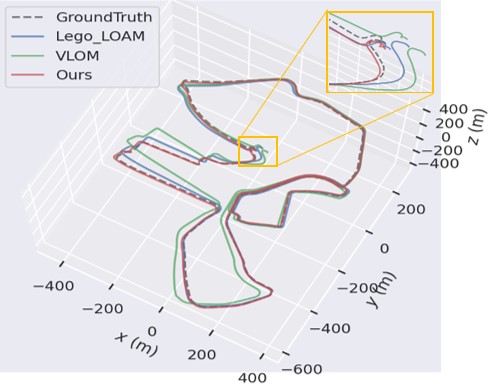}
		\end{minipage}
	}
    \subfigure[Utbm5]
    {
    	\begin{minipage}[b]{.46\linewidth}
    		\centering
    		\includegraphics[scale=0.55]{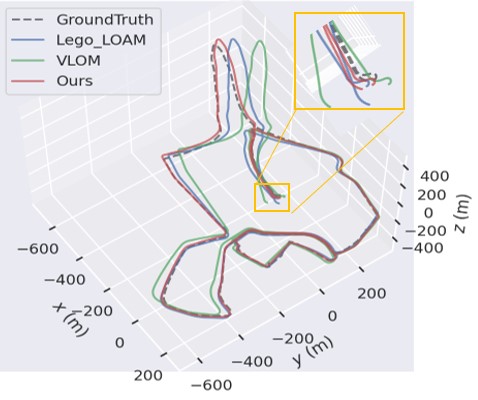}
    	\end{minipage}
    }
	\caption{Trajectory comparison on the UTBM dataset.}
	\label{fig:utbm}
\end{figure}

\subsection{Evaluation on Public Datasets}
In this letter, 11 sequences are selected from two public datasets to evaluate the accuracy of our algorithm. These sequences were collected from a test vehicle using mechanical LiDAR and get well GPS data were obtained as ground truth.
The UTBM dataset \cite{b22} was collected in Montpellier, France. Most of these long sequence trajectories are similar, but the acquisition time and environmental conditions are different. The short sequences of the ULHK dataset \cite{b23} were collected during rush hour on an urban street in Hong Kong filled with moving objects such as cars and people. 

Since our system does not include loopclosure detection, Lego-LOAM loopclosure detection has been manually disabled while all other modules are enabled for a fair comparison. We present the RMSE of the mapping ATE in Table \ref{public dataset}. Compared to other methods, our method achieves the best performance on all 11 sequences. In sequences 1-5 of the UTBM dataset, the error of our method is only half that of the other two methods. On the ULHK dataset, the sequence length is relatively short, so our accuracy has improved, but the extent is not particularly obvious.

In Fig. \ref{fig:utbm} we show the trajectories of the UTBM dataset sequences 2,3,4,and 5, including cloudy days, nights, snow and other conditions. The start and end points of the Utbm trajectories are shown enlarged. The end point of the red trajectory obtained by our algorithm almost coincides with the starting point and the whole process is closer to the ground truth.

\subsection{Evaluation of BA Odometry on Campus Datasets}
To validate the improvement of LMBAO in odometry, we performed ablation experiments on landmark map maintenance and marginalization using own campus dataset.

\begin{figure}[tp]
	\centering
	\includegraphics[width=0.99\linewidth]{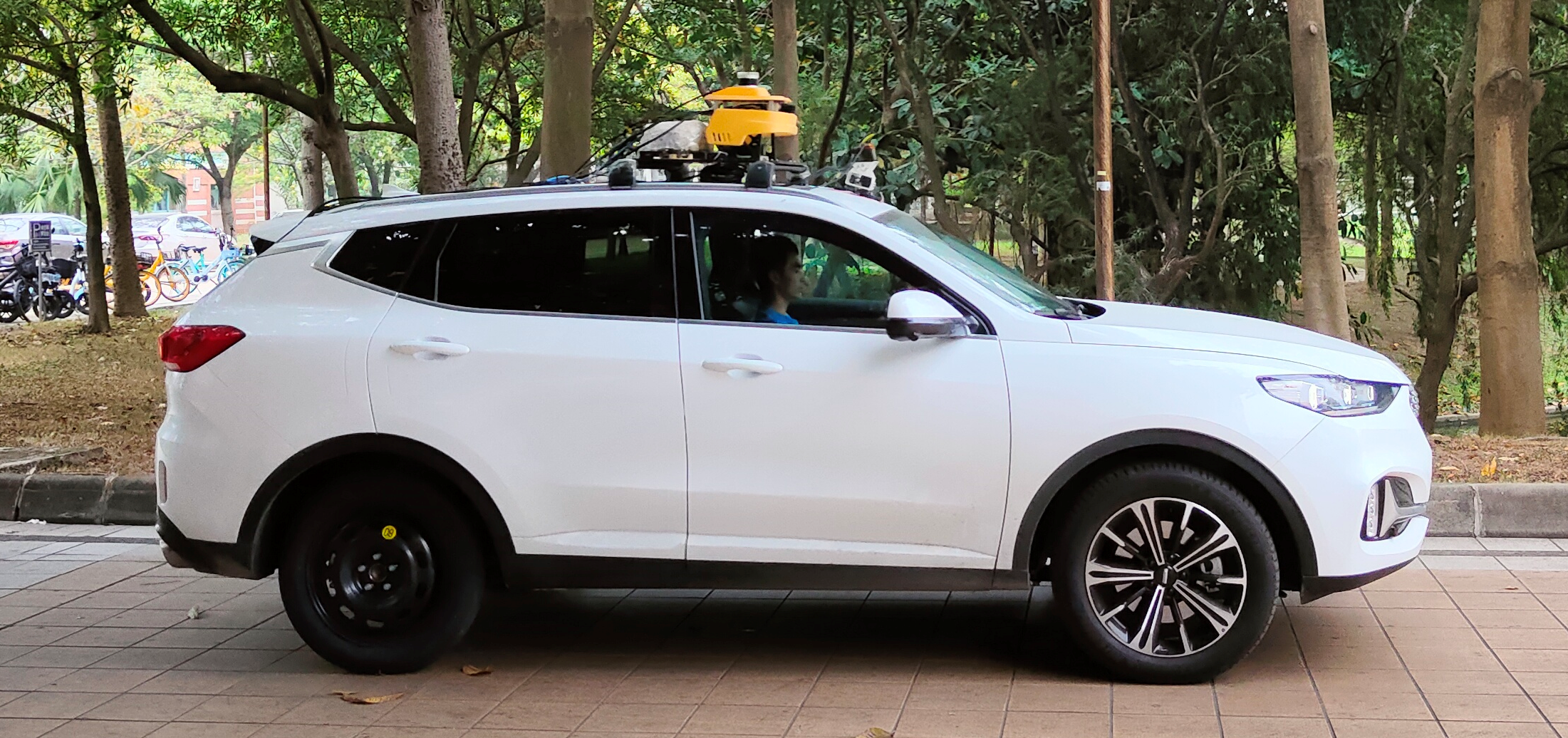}
	\caption{The sensor suite used in this letter.}
	\label{fig:carvv6}
\end{figure}

\begin{figure}
	\centering
	\includegraphics[width=0.99\linewidth]{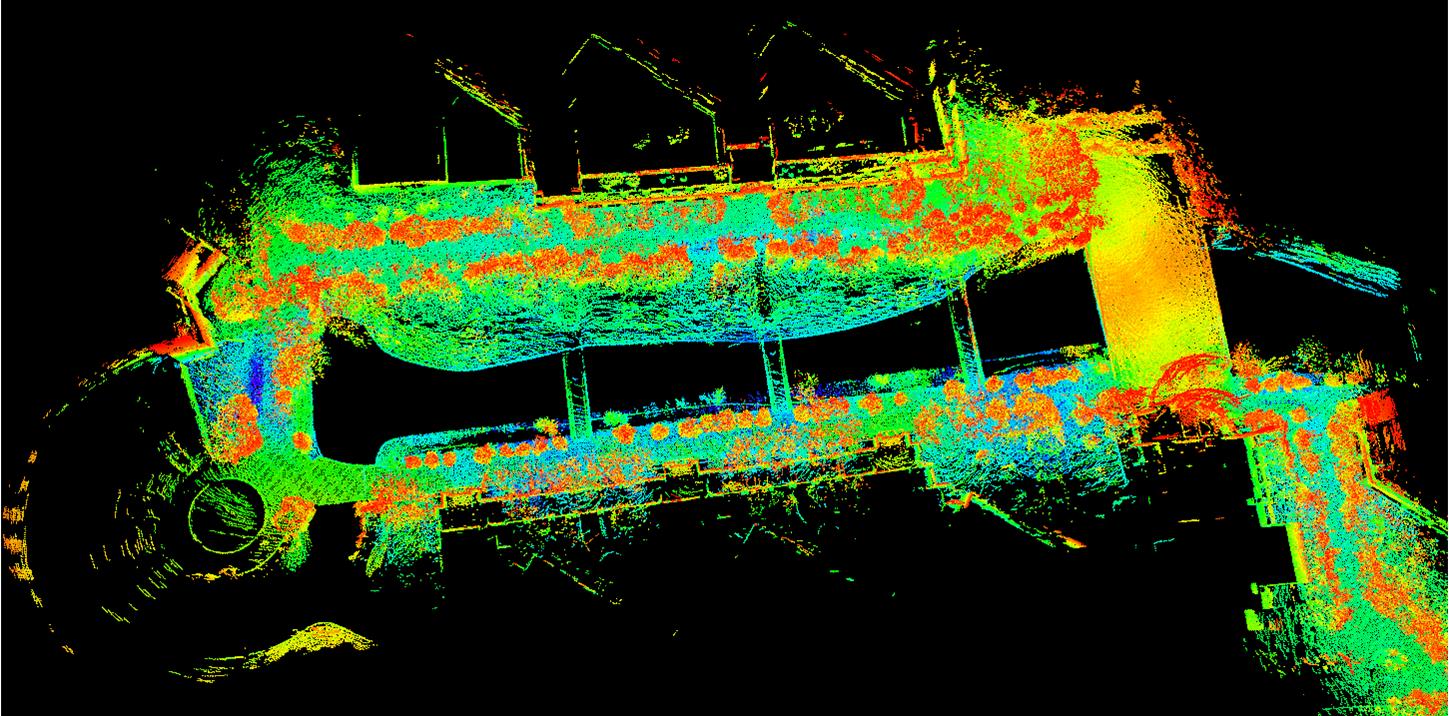}
	\caption[Map]{The point cloud map of river loop created by our method.}
	\label{fig:river}
\end{figure}

\textbf{Own Campus Dataset:} 
Three squences were collected on the eastern campus of Sun Yat-sen University at an average driving speed of 25 km/h and were named the library loop, river loop, and college loop. The sensor platform is shown in Fig. \ref{fig:carvv6}, and consists of an Ouster-64 LiDAR and a high-precision Newton-M2 navigation device to obtain the ground truth. The Newton-M2, in combination with a differential GNSS, can achieve an accuracy of 2 cm for position and an error of approximately $0.1\degree$ for attitude.
As a simple sequence, the college loop is seen near buildings with obvious features. The library loop surrounds the entire teaching area with many trees, including through open streets, making it a relatively difficult sequence. In the middle, the river loop goes around the river, with the riverbank on one side and buildings on the other, as shown in Fig. \ref{fig:river}.

\textbf{Variant LMBAO-LM:}
We develope a VLOM variant where only marginalization is added to the sliding window without our landmark map maintenance, named LMBAO-LM. The length of the sliding window is fixed at 4, as in LMBAO, and the last 8 scans outside the window are marginalized to contain a similar number of landmarks as in LMBAO.

\textbf{Accuracy Results:}
We evaluate the odometry modules of Lego-LOAM, VLOM, LMBAO-LM and LMBAO on 3 campus sequences. The RMSEs of odometry ATE are shown in Table \ref{Ours dataset}. 
The results of these three sequences from LMBAO all exceed the results from other three methods. 
Compared to VLOM, LMBAO-LM has more previous landmarks due to marginalization, which improves accuracy and significantly increases computational efficiency (in Table III).
At the same computational efficiency, LMBAO has significantly higher accuracy than LMBAO-LM due to the map maintenance strategy, which selects accurate and sufficient constraints for BA odometry from all prior landmark.


\begin{table}[t]
	\renewcommand\arraystretch{1.2}
	\caption{RMSE of the Odometry APE $(m)$ on the Campus Datasets}
	\centering
	\begin{tabular}{c|cccc}
		\hline
		Dataset         &library &river  & 	college \\
		\hline
		Length(km) & 2.928  &1.019 & 0.664\\
		\hline\hline
     	LegoLOAM-odom	&92.331 &14.254   & 6.4216   \\
		
		VLOM-odom& 25.873 &3.500 &1.0086 \\
		
	LMBAO-LM-odom&21.661 &  3.437 & 0.9610 \\
	Ours-odom &\textbf{\emph{13.098}} &\textbf{\emph{3.199}}& \textbf{\emph{0.857}}\\
		\hline
	\end{tabular}
	\label{Ours dataset}
\end{table}

\begin{figure}[tp]
	\centering
	\subfigure[Odometry trajectory comparison]
	{
		\begin{minipage}[b]{.7\linewidth}
			\centering
			\includegraphics[scale=0.14]{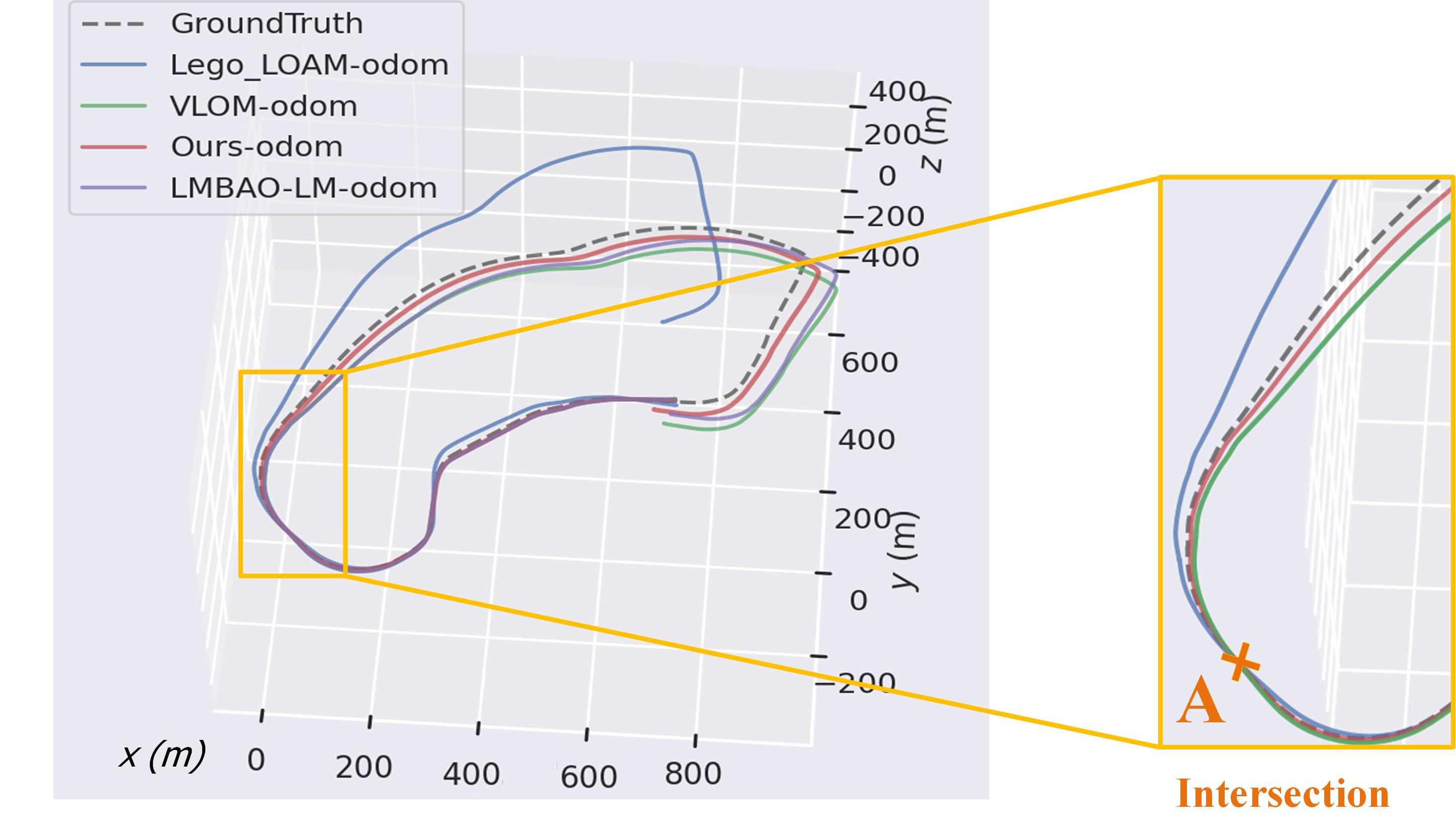}
		\end{minipage}
	}
    \subfigure[The top view at intersection A. ]
    {
    	\begin{minipage}[b]{.46\linewidth}
    		\centering
    		\includegraphics[scale=0.56]{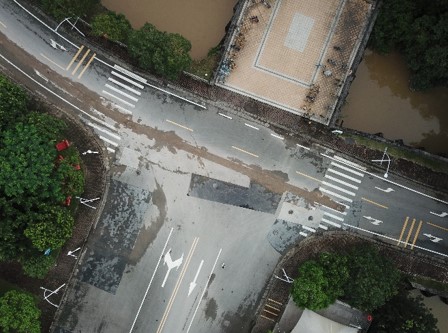}
    	\end{minipage}
    }
	\subfigure[Landmark map of VLOM.   (8 scans inside window.)]
	{
		\begin{minipage}[b]{.46\linewidth}
			\centering
			\includegraphics[scale=0.11]{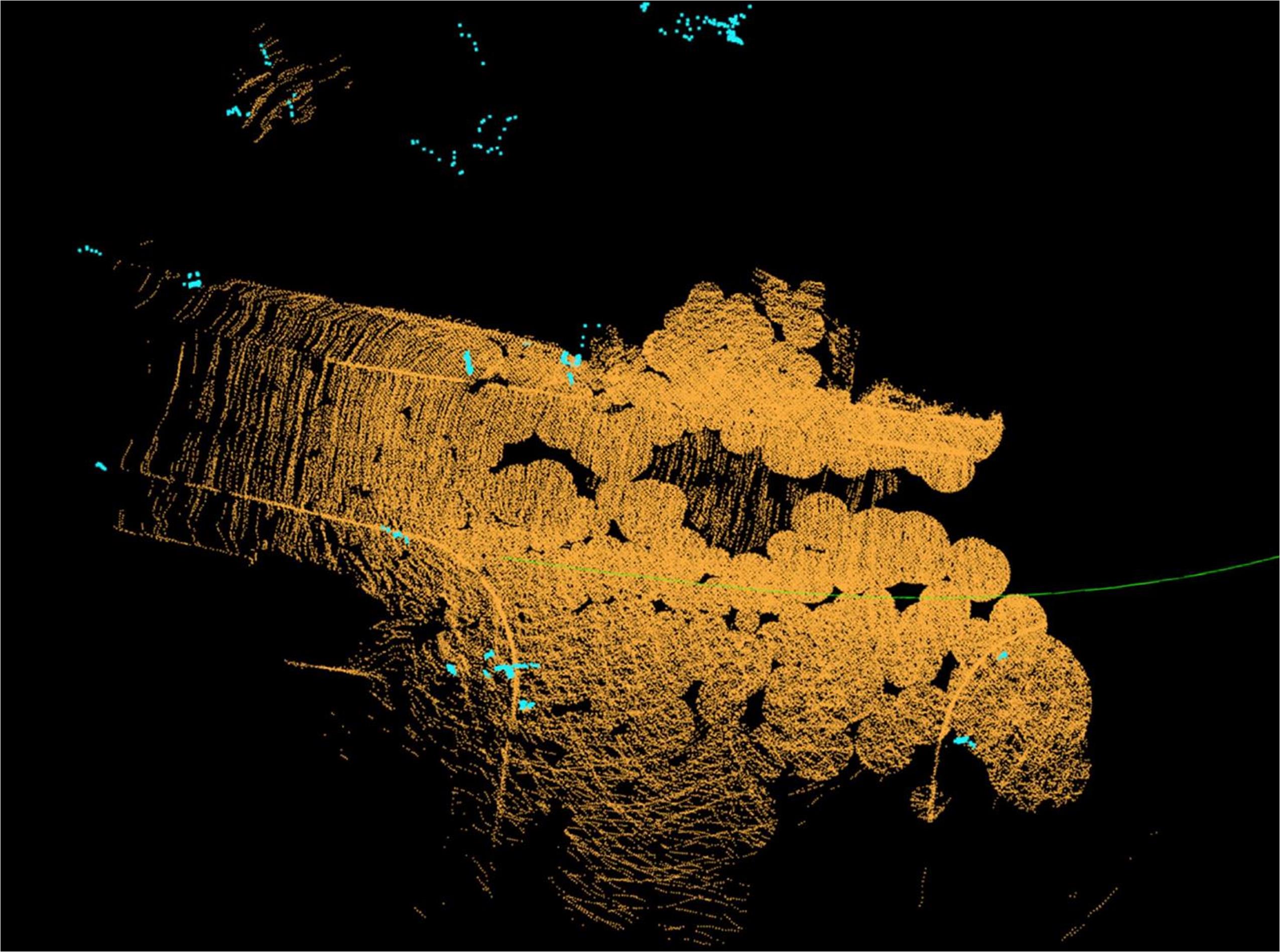}
		\end{minipage}
	}
	\subfigure[Landmark map of LMBAO-LM. (4 scans inside window, 8 scans ouside the window.)]
   {
	   \begin{minipage}[b]{.46\linewidth}
		\centering
		\includegraphics[scale=0.11]{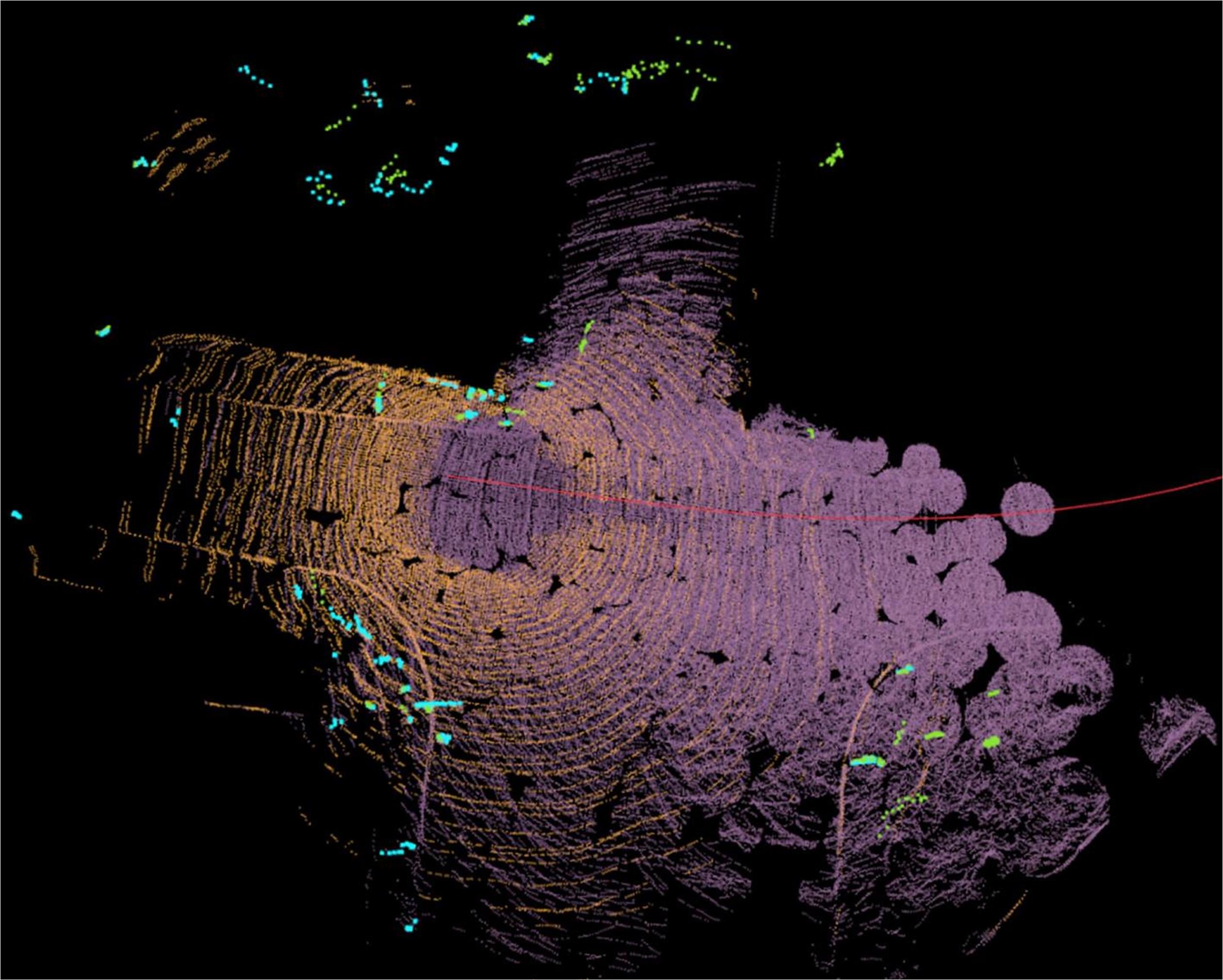}
	\end{minipage}
    }
	\subfigure[Landmark map of our LMBAO.   (4 scans inside window and some seleced priors outside window.)]
	{
		\begin{minipage}[b]{.46\linewidth}
			\centering
			\includegraphics[scale=0.11]{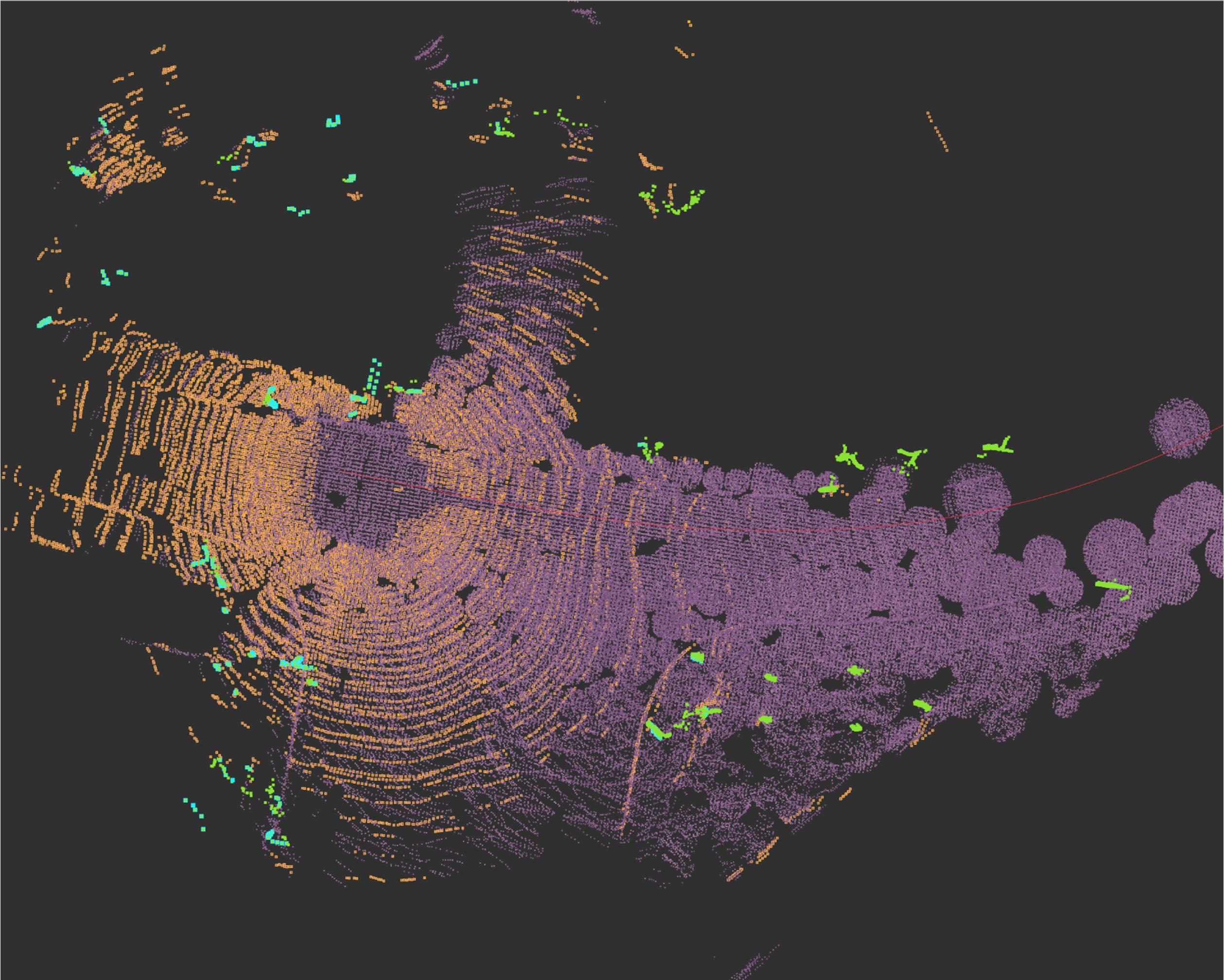}
		\end{minipage}
	}
	\caption{The odometry trajectory and landmark map of the library loop. In landmark map, planes and lines inside sliding widow are colored yellow and lightblue, and the marginal part outside are colored purple and green.}
	\label{fig:library}
\end{figure}

\textbf{Analysis on Intersection A:}
The odometry trajectories of the library loop shown in Fig. \ref{fig:library}(a), it is clear that the trajectories of Lego-LOAM and VLOM are significantly offset after passing the small intersection A, while LMBAO matches well with the ground truth. 
The landmark map created by of VLOM in Fig. \ref{fig:library}(c) is controlled by the sliding window and has few global feature - landmarks. 
Fig. \ref{fig:library}(d) of LMBAO-LM has more landmarks as marginal part but its prior constraints are still insufficient. In contrast, the landmark map generated by our LMBAO in \ref{fig:library}(e) preserves finer structural details of the environment, such as the light pole at the corner and the ground at the intersection. This demonstrates the effectiveness of our map maintenance and marginalization strategy when passing feature ambiguous areas, improving the robustness of BA odometry. 

\begin{table}[tp]
	\renewcommand\arraystretch{1.2}
	\caption{Computational Time $(ms)$ of Odometry}
	\centering 
	\begin{tabular}{c|ccc}
		\hline  
		Dataset & Utbm5 & Ulhk2 & Campus-library\\
		\hline
		VLOM  & 58.898 $\pm$ 6.064 &37.095 $\pm$ 3.793 &  42.011 $\pm$ 4.661  \\
	   LMBAO-LM  & 40.149 $\pm$ 5.987& 31.350 $\pm$ 3.008 & \textbf{41.382 $\pm$ 4.370}  \\
		Ours  & \textbf{36.312 $\pm$ 5.485}   & \textbf{29.647 $\pm$ 4.437} & 41.539 $\pm$ 4.518 \\
		
		\hline
	\end{tabular}
	\label{Runtime}
\end{table}

\subsection{Evaluation of Runtime}
To verify the real-time performance of the algorithm, we calculate the average runtime of the three modules processing a LiDAR frame in our LMBAO. The runtime performance of odometry is analysed using the longest sequence in the three datasets and the results are shown in Table \ref{Runtime}. The feature extraction step is similar to VLOM, its runtime is almost 40 ms. The mapping step is taken from Lego- LOAM, and its runtime is 50 ms, the same as the original. The propagation of the three modules in our whole pipeline is less than 100 ms, so we are able to run the LiDAR at a frequency of 10 Hz in real time.

Table \ref{Runtime} shows that the time requirement of LMBAO-LM is shorter than that of VLOM. Then, with the same number of landmarks as LMBAO-LM, LMBAO has the same computational cost, so their odometry runtimes are similar. Using marginalization to keep landmarks outside the sliding window and reducing the length of the sliding window greatly improve computational efficiency.

\section{CONCLUSIONS}
In this letter, we present a BA LiDAR odometry with active landmark map maintenance (LMBAO) that includes landmark association and landmark deletion.
We first describe the construction of the landmark map for BA, which is separated from the sliding window to fully exploit the prior constraints.
Then we performed the marginalization in combination with the landmark map in BA odometry, which greatly simplifies the computation by the covariance matrix.
Compared to other SLAM pipelines, errors in our pipeline are eliminated as early as possible, favoring subsequent optimization and even feature extraction, which works with our odometry in the same thread.
Furthermore, experiments on outdoor driving datasets demonstrate that LMBAO improves the robustness of BA odometry. In the comparison of state-of-the-art LiDAR SLAM algorithms, including Lego-LOAM and VLOM, LMBAO achieves better performance in terms of accuracy and efficiency.

\addtolength{\textheight}{-12cm}   









\bibliographystyle{IEEEtran}
\bibliography{References}

\end{document}